\documentclass[lettersize,journal]{IEEEtran}
\usepackage{amsmath,amsfonts}
\usepackage{algorithmic}
\usepackage{algorithm}
\usepackage{array}
\usepackage[caption=false,font=normalsize,labelfont=sf,textfont=sf]{subfig}
\usepackage{textcomp}
\usepackage{stfloats}
\usepackage{url}
\usepackage{verbatim}
\usepackage{graphicx}
\usepackage{cite}
\usepackage{bbm}
\usepackage{booktabs}
\usepackage{colortbl}
\usepackage{enumitem}
\usepackage{float}
\usepackage{hyperref}
\usepackage{multirow}
\usepackage{multicol}
\usepackage{tcolorbox}
\usepackage{xspace}
\newcommand{\eg}{e.g.\xspace}
\newcommand{\ie}{i.e.\xspace}
\definecolor{mygray}{gray}{.9}

\begin{document}

\title{mR$^2$AG: Multimodal Retrieval-Reflection-Augmented Generation for Knowledge-Based VQA}

\author{
    Tao Zhang$^*$, Ziqi Zhang$^*$, Zongyang Ma, Yuxin Chen, Zhongang Qi, Chunfeng Yuan, Bing Li$^\dagger$, Junfu Pu, Yuxuan Zhao, Zehua Xie, Jin Ma, Ying Shan, and Weiming Hu~\IEEEmembership{Senior Member,~IEEE}
    \thanks{$^*$ Tao Zhang and Ziqi Zhang contributed equally to this work.}
    \thanks{$^\dagger$ Bing Li is the corresponding author.}
    \thanks{Tao Zhang, Ziqi Zhang, Zongyang Ma, Chunfeng Yuan, Bing Li, and Weiming Hu are with the State Key Laboratory of Multimodal Artificial Intelligence Systems, Institute of Automation, Chinese Academy of Sciences, Beijing, 100190, China (e-mail: zhangtao2023@ia.ac.cn; zhangziqi2017@ia.ac.cn; mazongyang2020@ia.ac.cn; cfyuan@nlpr.ia.ac.cn; bli@nlpr.ia.ac.cn; wmhu@nlpr.ia.ac.cn).}
    \thanks{Tao Zhang and Weiming Hu are with the School of Artificial Intelligence, University of Chinese Academy of Sciences, Beijing, 100190, China.}
    \thanks{Tao Zhang, Ziqi Zhang, Zongyang Ma, Chunfeng Yuan, Bing Li, and Weiming Hu are with the Beijing Key Laboratory of Super Intelligent Security of Multi-Modal Information.}
    \thanks{Tao Zhang, Yuxin Chen, Junfu Pu, Yuxuan Zhao, Zehua Xie, Jin Ma, and Ying Shan are with Tencent Inc., Beijing, 100193 China (e-mail: chenyux53@163.com; jevinpu@tencent.com; martyzhao@tencent.com;  zehuaxie@tencent.com; daniellwang@tencent.com; yingsshan@tencent.com).}
    \thanks{Bing Li is with PeopleAl, Inc., Beijing, 100190, China.}
    \thanks{Zhongang Qi is with the Huawei Noah’s Ark Laboratory, Beijing, 100085 China (e-mail: zhongangqi@gmail.com). Work done during Zhongang Qi's tenure at Tencent PCG ARC Lab.}
    \thanks{Weiming Hu is with the School of Information Science and Technology, ShanghaiTech University, Shanghai, 201210, China.}
    \thanks{This paper has supplementary downloadable material available at http://ieeexplore.ieee.org, provided by the authors. The material includes prompt templates used for dataset annotation and inference, and additional qualitative visualization results.}
}

\markboth{IEEE Transactions on Multimedia}%
{Zhang \MakeLowercase{\textit{et al.}}: mR$^2$AG: Multimodal Retrieval-Reflection-Augmented Generation for Knowledge-Based VQA}


\maketitle

\begin{abstract}
Advanced Multimodal Large Language Models (MLLMs) struggle with recent Knowledge-based Visual Question Answering (VQA) tasks, such as INFOSEEK and Encyclopedic-VQA, due to their limited and frozen knowledge scope, often leading to ambiguous and inaccurate responses. Thus, multimodal Retrieval-Augmented Generation (mRAG) is naturally introduced to provide MLLMs with comprehensive and up-to-date knowledge, effectively expanding the knowledge scope. However, current mRAG methods have inherent drawbacks, including: 1) Performing retrieval even when external knowledge is not needed. 2) Lacking of identification of evidence that supports the query. 3) Increasing model complexity due to additional information filtering modules or rules. To address these shortcomings, we propose a novel generalized framework called \textbf{m}ultimodal \textbf{R}etrieval-\textbf{R}eflection-\textbf{A}ugmented \textbf{G}eneration (mR$^2$AG), which achieves adaptive retrieval and useful information localization to enable answers through two easy-to-implement reflection operations, preventing high model complexity. In mR$^2$AG, Retrieval-Reflection is designed to distinguish different user queries and avoids redundant retrieval calls, and Relevance-Reflection is introduced to guide the MLLM in locating beneficial evidence of the retrieved content and generating answers accordingly. In addition, mR$^2$AG can be integrated into any well-trained MLLM with efficient fine-tuning on the proposed mR$^2$AG Instruction-Tuning dataset (mR$^2$AG-IT). mR$^2$AG significantly outperforms state-of-the-art MLLMs (\eg, GPT-4o) and mRAG-based MLLMs on INFOSEEK and Encyclopedic-VQA, while maintaining the exceptional capabilities of base MLLMs across a wide range of Visual-dependent tasks. 
\end{abstract}

\begin{IEEEkeywords}
Multimodal Large Language Models, Retrieval-Augmented Generation, Knowledge-based Visual Question Answering, Instruction Tuning
\end{IEEEkeywords}

\section{Introduction}

\begin{figure}
  \centering
  \includegraphics[width=\columnwidth]{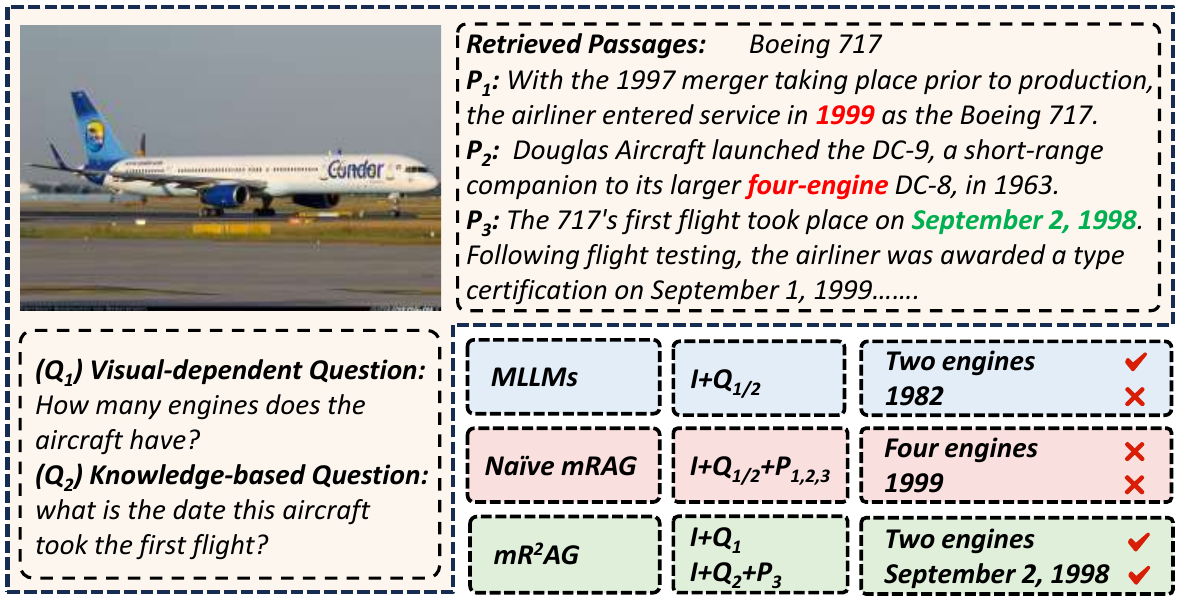}
  \caption{
  Comparisons of different methods on Visual-dependent and Knowledge-based VQA tasks:
  1) Typical MLLMs use the image $I$ and question $Q$ as inputs, offering limited support for Knowledge-based questions.
  2) Naive mRAG use $I$, $Q$, and retrieved content $P_{1,2,3}$ as inputs in all cases, inevitably introducing irrelevant noise.     
  3) mR$^2$AG adaptively determines the necessity of retrieval and effectively locates the useful context, \ie, $P_3$ for $Q_2$.
  }
  \label{fig:1}
\end{figure}

\IEEEPARstart{T}he rapid development of Multimodal Large Language Models (MLLMs) \cite{alayrac2022flamingo, li2023blip, liu2024improved,  achiam2023gpt, reid2024gemini, zhu2024mipha, dong2024internlm, liu2023visual, wu2024deepseek, bai2025qwen2, hong2024cogvlm2} enables them to excel in Visual-dependent VQA tasks that rely solely on visual content or common sense, such as VQAv2 \cite{goyal2017making}, GQA \cite{hudson2019gqa} and TextVQA \cite{singh2019towards}. However, recently proposed Knowledge-based VQA tasks like INFOSEEK \cite{chen2023can} and Encyclopedic-VQA (E-VQA)\cite{mensink2023encyclopedic}, introduce fine-grained visual entities and focus on encyclopedic knowledge, posing significant new challenges for existing MLLMs. As shown in $Q_2$ of Fig.~\ref{fig:1}, when queried about the first flight date of an airplane, typical MLLMs tend to provide inaccurate and overly general responses due to their limited and frozen knowledge scope. 

To obtain accurate and specific answers, some works \cite{hu2024avis, jian2024large, caffagni2024wiki, yan2024echosight, qiu2024snapntell, qi2024rora} introduce multimodal Retrieval-Augmented Generation (mRAG) that leverages external knowledge bases. They first use retrievers to find query-related information, then input them into the MLLM for answer generation. Nevertheless, several challenges remain: 
\textbf{1) Indiscriminate use of retrieval.} 
Visual-dependent questions usually do not require external knowledge, and conducting retrieval for them may introduce noise and lead to wrong answers, such as the response of Naive mRAG to $Q_1$ in Fig.~\ref{fig:1}.
\textbf{2) Lack of explicit evidence localization.} 
When encountering questions beyond one's knowledge, humans first search for relevant information and then obtain reliable answers by locating direct evidence, \eg, $P_3$ of $P$ in Fig.~\ref{fig:1}. 
However, current methods \cite{caffagni2024wiki, yan2024echosight, qiu2024snapntell} input retrieved text into the model to directly generate answers, lacking an explicit evidence localization process, making it difficult to determine whether the model can effectively utilize the useful retrieved information.
\textbf{3) High model complexity.} To improve performance, some methods \cite{caffagni2024wiki, hu2024avis, yan2024echosight, qiu2024snapntell, jian2024large} introduce complex rules or even external models to calculate query-passage correlations to filter retrieved content.

To address the above challenges, we propose a novel generalized framework called \textbf{m}ultimodal \textbf{R}etrieval-\textbf{R}eflection-\textbf{A}ugmented \textbf{G}eneration (mR$^2$AG), designed to seamlessly leverage the inherent instruction-following, multimodal understanding and reasoning abilities in MLLMs to enhance Knowledge-based question answering. For clarifying the necessity of retrieval, mR$^2$AG introduces \textbf{Retrieval-Reflection} to determine whether the user's query is Knowledge-based or Visual-dependent for adaptive retrieval, thus expanding MLLMs' knowledge scope while maintaining the original performance. Furthermore, mR$^2$AG imitates human behavior to implement \textbf{Relevance-Reflection}, which guides the model to explicitly assess the evidence parts in all retrieved information, and generates accurate responses based on the identified evidence. The implementation of the proposed reflection operations involves only modifications to the MLLMs' vocabulary, without introducing any additional modules or computational strategies to destroy the original structure of the models, thereby effectively coupling with the MLLMs' existing capabilities.

To quickly integrate mR$^2$AG into pre-trained MLLMs, we provide a corresponding \textbf{mR$^2$AG} \textbf{I}nstruction \textbf{T}uning dataset (mR$^2$AG-IT) specifically constructed for Knowledge-based VQA tasks through an automated annotation pipeline. Specifically, mR$^2$AG-IT annotates the evidence paragraphs within Wikipedia articles that explicitly support user queries. Comprehensive comparisons on Knowledge-based VQA datasets show that our approach significantly outperforms existing methods. When using LLaVA-v1.5-7B \cite{liu2024improved} as the base MLLM, applying mR$^2$AG not only achieves performance gains of 10.6\% and 15.5\% over the previous SOTAs on the INFOSEEK$_{\text{Human}}$ and INFOSEEK$_{\text{Wikidata}}$ test sets, but also surpasses SOTAs on the Encyclopedic-VQA (E-VQA) test set by 2.5\% on single-hop questions and 18.2\% on multi-answer questions. Moreover, our method retains the excellent capabilities of the base MLLM on Visual-dependent VQA tasks, demonstrating performance comparable to LLaVA-v1.5-7B across various benchmarks.

The contributions of this work are summarized as follows:
\begin{itemize}
    \item We propose an advanced multimodal RAG framework \textbf{mR$^2$AG}, which only uses two reflection operations to stimulate MLLMs to implement retrieval invocation, evidence content identification, and answer generation.
    
    \item We provide the \textbf{mR$^2$AG-IT} dataset, which is designed to facilitate the rapid adaptation of MLLMs to Knowledge-based VQA and serves as a supplement to general visual instruction tuning datasets.

    \item mR$^2$AG, with its concise and effective design, can be seamlessly integrated with common MLLMs. It significantly outperforms existing mRAG methods in answering Knowledge-based queries while maintaining the base MLLM's capabilities in Visual-dependent tasks.
\end{itemize}

\section{Related Work}
\label{sec:Related_work}

\textbf{Knowledge-based VQA.}
Several works \cite{mensink2023encyclopedic, chen2023can} focus on Knowledge-based VQA and construct corresponding datasets.  Early OK-VQA \cite{marino2019ok} and its variant A-OKVQA \cite{schwenk2022okvqa} emphasize the significance of knowledge in VQA but primarily focus on commonsense. ViQuAE \cite{lerner2022viquae} introduces a wide range of entity types and tests fine-grained knowledge related to named entities. However, certain questions in ViQuAE \cite{lerner2022viquae} can be answered without viewing the visual content. INFOSEEK \cite{chen2023can} and E-VQA \cite{mensink2023encyclopedic} address this by designing questions that force the model to examine the image for the correct answer. INFOSEEK categorizes these questions into two splits: Unseen Entity and Unseen Question. E-VQA introduces multiple question types , including single-hop, multi-answer, and two-hop questions. These datasets cover a wide range of Wikipedia entities, focusing on fine-grained knowledge related to these entities, and additionally provide a multimodal Wikipedia knowledge base as an external source of knowledge.

\textbf{Multimodal Large Language Models.}
Advancements in large language models (LLMs) \cite{vaswani2017attention, brown2020language, ouyang2022training, chowdhery2023palm, touvron2023llama, chiang2023vicuna, chung2024scaling, anil2023palm, yang2024qwen2, dao2024transformers, guo2025deepseek}, particularly through architectural innovations and increased training scale, expand their knowledge scope and enhance their language understanding and reasoning abilities. These improvements enable multimodal large language models (MLLMs) \cite{alayrac2022flamingo, li2023blip, liu2024improved, liu2023visual, achiam2023gpt, reid2024gemini, zhu2024mipha, dong2024internlm, wu2024deepseek, bai2025qwen2, hong2024cogvlm2} to make significant progress by using LLMs as core cognitive component and incorporating cross-modal alignment techniques. Typical MLLMs, \eg, LLaVA \cite{liu2023visual}, widely adopt a two-stage pre-training and fine-tuning paradigm, achieving impressive capabilities across various multimodal tasks \cite{singh2019towards, li2023seed, yue2023mmmu, li2023evaluating, 10004507, 10643329, 10119213, 10374273}. MLLMs perform well in understanding human queries \cite{liu2024improved}, handling purely visual VQA tasks \cite{goyal2017making, hudson2019gqa, 9488320, 9206083, 10477535, 10643329}, and addressing commonsense VQA tasks \cite{marino2019ok}. However, they struggle with Knowledge-based VQA tasks that involve fine-grained knowledge of specific visual entities, as shown by their performance on INFOSEEK \cite{chen2023can} and E-VQA \cite{mensink2023encyclopedic}.

\textbf{Retrieval-Augmented Generation.}
RAG is widely used in LLMs to address challenges like hallucinations \cite{zhang2023siren}, non-renewable knowledge, and opaque reasoning \cite{kandpal2023large, gao2023retrieval}. It combines LLMs' inherent knowledge with dynamic external knowledge, offering a solution for knowledge-intensive tasks \cite{lewis2020retrieval, mao2020generation}. This technique is gaining traction within the MLLMs domain \cite{hu2024avis, jian2024large, caffagni2024wiki, yan2024echosight, qiu2024snapntell, qi2024rora, 10404060, 10163831}. For example, Wiki-LLaVA \cite{caffagni2024wiki} retrieves Wikipedia articles from the input image and uses Contriever \cite{gautier2022unsupervised} to select relevant passages. EchoSight \cite{yan2024echosight} introduces a fine-tuned Q-Former \cite{li2023blip} architecture as a reranker, filtering the retrieved content based on both the input image and question. RORA-VLM \cite{qi2024rora} introduces a noise-resistant RAG method, implicitly guiding the model to focus on retrieved content more relevant to the query. However, it still lacks clear identification of the evidence source. These methods mainly rely on external modules to filter retrieved information, ignoring the inherent multimodal understanding and evidence discrimination capabilities of MLLMs. Inspired by SELF-RAG \cite{asai2023self}, we propose the mR$^2$AG framework, which leverages MLLMs to independently localize the query-relevant evidence within the retrieved content, eliminating the need for additional modules or complex strategies.

\begin{figure*}[ht]
  \centering
  \includegraphics[width=\textwidth]{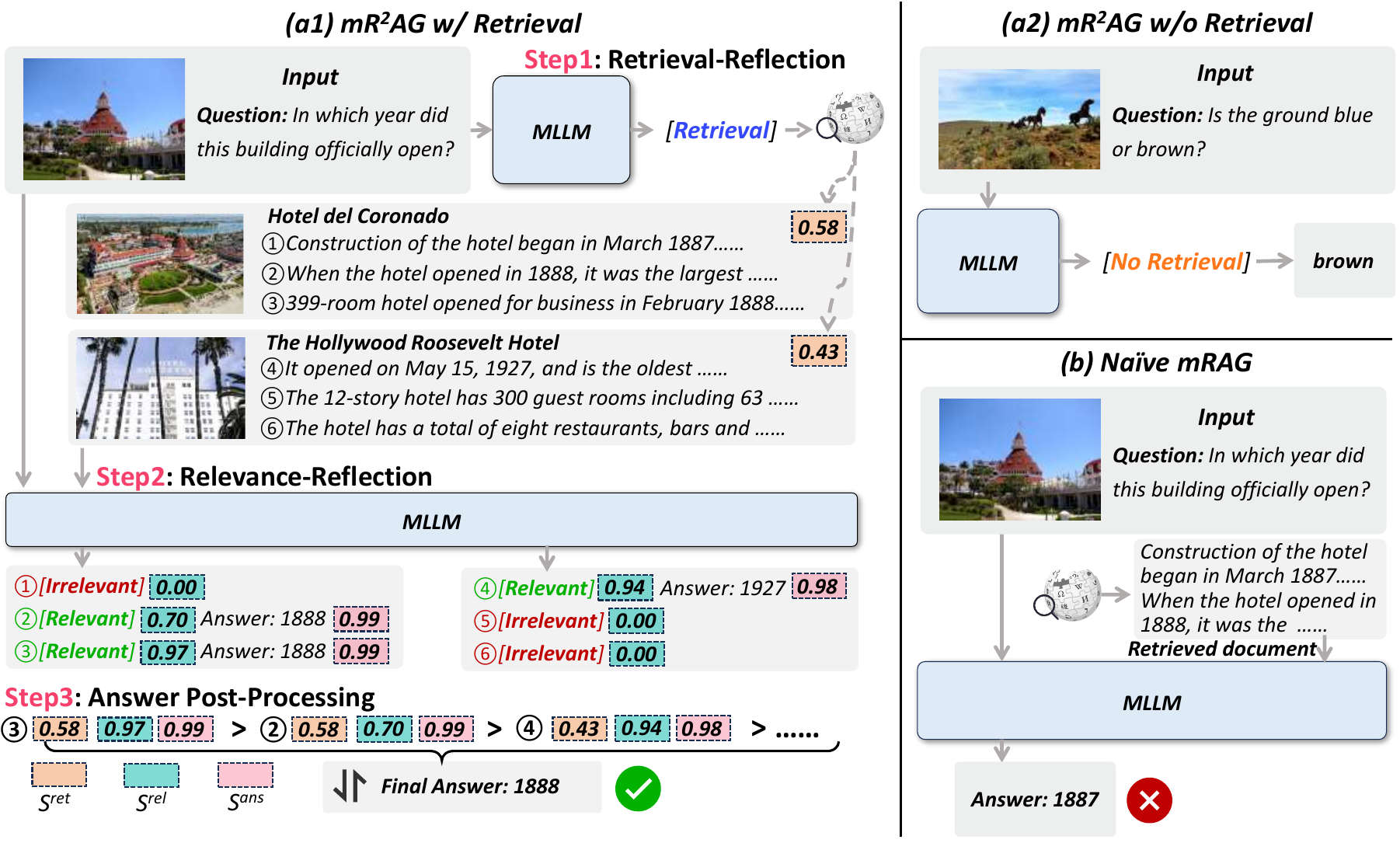}
  \caption{
  Overview of the mR$^2$AG framework.
  (a1) mR$^2$AG w/ Retrieval: This process includes:
  a) Retrieval-Reflection for determining the necessity of retrieval;
  b) Relevance-Reflection for identifying evidence passages;
  c) Post-processing multiple potential answers.
  (a2) mR$^2$AG w/o Retrieval: The generation process when retrieval is unnecessary.
  (b) Naïve mRAG: A baseline method without reflection.
  }
  \label{fig:2}
\end{figure*}

\section{Methodology}
\label{sec:Methodology}

Knowledge-based VQA takes an image-question pair $(I, Q)$ as input and is supported by an accessible knowledge base.
As shown in Fig.~\ref{fig:2}, naive mRAG first retrieves the top-$N$ articles most relevant to the input from the knowledge base, denoted as $\hat{P} = \{\hat{P_i}\}_{i=1}^{N}$, and then feeds the concatenation of $\hat{P}$ and the $(I, Q)$ into the MLLM to directly generate the response $y^{\mathrm{ans}}$.
In contrast, mR$^2$AG proposes two novel reflection operations to decouple the generation process into three steps: 
(1) Performing Retrieval-Reflection to determine whether retrieval is needed.
(2) Performing Relevance-Reflection to identify evidence passages and generate answers  accordingly.
(3) Post-processing multiple potential answers.
The presentation of mR$^2$AG is organized as follows:
Section \ref{sec:generation} introduces the mR$^2$AG method.
Section \ref{sec:training} describes the training of mR$^2$AG. 
Section \ref{sec:dataset} introduces the mR$^2$AG-IT dataset for fine-tuning.

\subsection{\texorpdfstring{mR$^2$AG Method}{mR2AG Method}}
\label{sec:generation}

\subsubsection{Retrieval-Reflection}
\label{sec:retrieval}
User queries can be divided into Visual-dependent and Knowledge-based according to the input $(I, Q)$. 
As shown in Fig.~\ref{fig:2}, the question in case $(a1)$ requires external knowledge for a confident answer, while the question in case $(a2)$ can be answered entirely relying on visual content. Introducing external knowledge in the latter case may bring undesirable noise.
To guide the model in distinguishing between different queries, we define two special tokens: \textbf{[Retrieval]/[No Retrieval]}, to perform Retrieval-Reflection.
First, the model generates retrieval-reflection predictions $y^{\mathrm{ret}}$ based on the input $(I, Q)$:
\begin{equation}
    y^{\mathrm{ret}} = \textrm{MLLM}(I,Q).
\end{equation}
Depending on the different results of $y^{\mathrm{ret}}$, one of the following ways is executed:
\begin{itemize}[leftmargin=*]
    \item $y^{\mathrm{ret}}$ = \textbf{[No Retrieval]}:
    The model determines that the question can be answered without external knowledge, and conditions on this token along with $(I, Q)$ to generate the answer $y^{\mathrm{ans}}$:
        \begin{equation}
            y^{\mathrm{ans}} = \textrm{MLLM}(I,Q,y^{\mathrm{ret}}=\textrm{[No Retrieval]}).
        \end{equation}
    \item $y^{\mathrm{ret}}$ = \textbf{[Retrieval]}: The model recognizes the need for external knowledge to answer the question and invokes retrievers to assist in further generation process.
\end{itemize}

We use English Wikipedia entries as the knowledge base, where the $k_{th}$ entry consists of a candidate image $\hat{I_k}$, title $\hat{T_k}$, and article $\hat{P_k}$.
mR$^2$AG combines cross-modal and uni-modal retrieval to select the most relevant Wikipedia entries to the query image $I$.
CLIP \cite{radford2021learning} is utilized to encode $I$, $\hat{I_k}$ and $\hat{T_k}$, and calculates the cosine similarity of $\mathrm{sim}(I, \hat{I_k})$ and $\mathrm{sim}(I, \hat{T_k})$. 
The overall retrieval score $S^{\mathrm{ret}}_k$ for the $k_{th}$ entry is the average of two cosine similarities:
\begin{equation}
\label{score:sim}
S^{\mathrm{ret}}_k =  \left( \mathrm{sim}(I, \hat{I_k}) + \mathrm{sim}(I, \hat{T_k}) \right) /2.
\end{equation}
The supposed result $\hat{P} = \{\hat{P_i}\}_{i=1}^{N}$ correspond to the articles of the top-$N$ entries with the highest retrieval scores.

\subsubsection{Relevance-Reflection}
\label{sec:assessment}

We divide each retrieved article $\hat{P}_{i}$ into multiple natural paragraphs.
To enable the model to determine whether each segmented paragraph $\hat{P}_{ij}$ contains evidence relevant to the question $Q$, we introduce two relevance-reflection tokens: \textbf{[Relevant]/[Irrelevant]}.
The model conditioned on the combination of $\hat{P}_{ij}$ and the query $(I, Q)$ generates the relevance-reflection prediction $y^{\mathrm{rel}}_{ij}$:
\begin{equation}
\label{score: rel}
    y^{\mathrm{rel}}_{ij} = \mathrm{MLLM}(I, Q, \hat{P}_{ij}).
\end{equation}
According to the result of $y^{\mathrm{rel}}_{ij}$, mR$^2$AG selects one of the following processes to perform:
\begin{itemize}[leftmargin=*]
    \item $y^{\mathrm{rel}}_{ij}$ = \textbf{[Irrelevant]:}   
    This indicates that the model perceives the $\hat{P}_{ij}$ as irrelevant to the query and lacking sufficient evidence, prompting the model to terminate the generation process and avoid producing unreliable answers.
    
    \item $y^{\mathrm{rel}}_{ij}$ = \textbf{[Relevant]:} The model considers $\hat{P_{ij}}$ relevant to the query, containing evidence beneficial for answer generation, and thus proceeds to generate the answer $y^{\mathrm{ans}}_{ij}$ based on $(I, Q, \hat{P}_{ij}, y^{\mathrm{rel}}_{ij})$:
    \begin{equation}
        y^{\textrm{ans}}_{ij} = \textrm{MLLM}(I, Q, \hat{P_{ij}}, y^{\textrm{rel}}_{ij} = \textrm{[Relevant]}).
    \end{equation}
\end{itemize}

\subsubsection{Answer Post-Processing}
\label{sec:post-processing}

Multiple evidence passages may exist in an article, leading to generate multiple candidate answers.
Therefore, post-processing is necessary to arrive at a single final answer.
Based on the retrieval-reflection-augmented generation process, we apply a hierarchical post-processing to rank the candidate answers by integrating scores at three levels:
\begin{itemize}[leftmargin=*]
    \item \textbf{Entry-Level.} The retrieval score in Equation.~\ref{score:sim} measures the similarity between the query image $I$ and the candidate Wikipedia entry, which serves as the Retrieval-Reflection score $S^{\mathrm{ret}}_{i}$  for the $i_{th}$ retrieved entry.

    \item \textbf{Passage-Level.} The probability of generating the \(\textrm{[Relevant]}\) Relevance-Reflection token quantifies the model's confidence in judging $\hat{P}_{ij}$ as evidence, which can be defined as the Relevance-Reflection score \(S^{\mathrm{rel}}_{ij}\):
    \begin{equation}
        S^{\mathrm{rel}}_{ij} = p_{\theta}\left( y^{\mathrm{rel}}_{ij} = \textrm{[Relevant]} \mid I, Q, \hat{P}_{ij} \right),
    \end{equation}
    where \(\theta\) represents the parameters of MLLM.
    
    \item \textbf{Answer-Level.} We calculate the probability of each token in the generated answer sequence and use the geometric mean to normalize the influence of sequence length variation, resulting in the answer confidence score \(S^{\mathrm{ans}}_{ij}\):
    \begin{equation}
    \label{score:ans}
        S^{\mathrm{ans}}_{ij} =  \sqrt[n]{p_{\theta}(y^{\textrm{ans}}_{ij})},
    \end{equation}
    where \(n\) represents the sequence length.
    This score reflects the model's confidence in generating answers based on the retrieved content and reflection tokens.
\end{itemize}
\textbf{Post-processing.}  The three levels of scores comprehensively consider each step in the answer generation process, evaluating the reliability of the candidate answers at the entry, passage, and answer levels, respectively. The effects of the three scores are integrated by calculating their product, which serves as the final criterion for ranking candidate answers. The model outputs the answer with the highest score based on this criterion.

\subsection{Training with Reflection Mechanism}
\label{sec:training}
During instruction tuning, we combine the common visual instruction tuning dataset, such as LLaVA-IT \cite{liu2024improved}, with the specifically designed mR$^2$AG-IT dataset:

\begin{itemize}[leftmargin=*]
    \item For each sample in LLaVA-IT, we set the Retrieval-Reflection token to [No Retrieval]. 
    The model is trained to answer the questions depends only on the visual content, and the training loss is formulated as:
    \begin{align}
        \mathcal{L}_1 &= -\mathbb{E}_{(I,Q,y^{ret},y^{\mathrm{ans}}) \sim \mathcal{D}_{\text{LLaVA-IT}}} \\
        & \log p_{\theta}(y^{\mathrm{ret}}= [\text{No Retrieval}], y^{\mathrm{ans}} \mid I,Q). \notag
    \end{align}

    \item For each sample in mR$^2$AG-IT, we set the Retrieval-Reflection token to [Retrieval]. 
    The model is trained to invoke retrievers, identify evidence passages, and generate accurate response. The training loss can be defined as:
    \begin{align}
        \mathcal{L}_2 = & -\mathbb{E}_{(I,Q,y^{ret},\hat{P}_{ij},y^{rel}_{ij},y^{\mathrm{ans}}) \sim \mathcal{D}_{\text{mR$^2$AG-IT}}} \\
        & \Big( \log p_{\theta}(y^{ret}=[\text{Retrieval}] |I,Q) \notag \\
        & + \mathbbm{1}\big[y^{rel}_{ij}=[\text{Relevant}]\big] \log p_{\theta}(y^{rel}_{ij}, y^{\mathrm{ans}} |I,Q,\hat{P}_{ij}) \notag \\
        & + \mathbbm{1}\big[y^{rel}_{ij}=[\text{Irrelevant}]\big]  \log p_{\theta}(y^{rel}_{ij} |I,Q,\hat{P}_{ij})\Big), \notag
    \end{align}
where the indicator function $\mathbbm{1}[\cdot]$ equals to 1 when the condition inside the parentheses is satisfied and 0 otherwise.
\end{itemize}

\subsection{\texorpdfstring{mR$^2$AG-IT Dataset}{mR2AG-IT Dataset}}
\label{sec:dataset}

For the Knowledge-based VQA tasks involving INFOSEEK \cite{chen2023can} and E-VQA \cite{mensink2023encyclopedic}, we propose an automated pipeline for annotating training data.
Each sample in these training datasets includes a \((I, Q, P_i, A)\) quadruple, where \(P_i\) is the ground truth Wikipedia article containing the answer source.

For INFOSEEK \cite{chen2023can}, the annotation process includes:
\begin{enumerate}[leftmargin=*]
    \item The Wikipedia article is segmented into natural paragraphs, with each paragraph \(P_{ij}\) preserving semantic independence and forming a new quadruple \((V, Q, P_{ij}, A)\).
    \item Each paragraph \(P_{ij}\) is evaluated by GPT-4 \cite{achiam2023gpt} with the question \(Q\) and answer \(A\) to determine if it serves as evidence.
    If \(P_{ij}\) is relevant to the query and supports answer generation, the Relevance-Reflection token is labeled as $y^{rel}_{ij}$ = [Relevant]; otherwise, $y^{rel}_{ij}$ = [Irrelevant].
    \item Due to the limited quantity of Wikipedia articles in the INFOSEEK training set, we additionally incorporate the Natural Questions (NQ) dataset \cite{kwiatkowski2019natural} as a supplement. This dataset comprises of real queries from the Google search engine, each with a long answer (when available) and a short answer as the response to the query. We consider the long answer as evidence for the query and the short answer as the response.
\end{enumerate}

For E-VQA \cite{mensink2023encyclopedic}, each sample is annotated with the evidence paragraphs from the ground truth Wikipedia article \(P_i\) that support the answer, eliminating additional steps to identify the evidence paragraphs. To ensure the precision of all training samples, we conduct strict string searches for filtering, ensuring answers appear only in evidence paragraphs.

\section{Experiments}
\label{sec:Experiments}

\begin{table*}[ht]
\caption{Main results of models on the INFOSEEK. $\dagger$ denotes our method and its variants with alternative designs.}
\label{tab:infoseek}
\centering
\footnotesize
\resizebox{\textwidth}{!}{%
\begin{tabular}{>{\arraybackslash}p{3.6cm}>{\arraybackslash}p{1.0cm}>{\raggedleft\arraybackslash}p{1.0cm}ccccccccc}
    \toprule
    \multirow{3}{*}{Model}  & \multirow{3}{*}{LLM} & \multirow{3}{*}{\#Params} & \multicolumn{3}{c}{INFOSEEK$_{\text{Wikidata}}$} & \multicolumn{3}{c}{INFOSEEK$_{\text{Human}}$} & \multicolumn{3}{c}{INFOSEEK$_{\text{Validation}}$}\\
                            &                      &                           & \begin{tabular}[c]{@{}c@{}}Unseen\\Question\end{tabular} & \begin{tabular}[c]{@{}c@{}}Unseen\\Entity\end{tabular} & Overall & \begin{tabular}[c]{@{}c@{}}Unseen\\Question\end{tabular} & \begin{tabular}[c]{@{}c@{}}Unseen\\Entity\end{tabular} & Overall & 
                            \begin{tabular}[c]{@{}c@{}}Unseen\\Question\end{tabular} & \begin{tabular}[c]{@{}c@{}}Unseen\\Entity\end{tabular} & Overall \\
    \midrule
    \multicolumn{12}{l}{\textbf{Without Knowledge}} \\
    LLaVA \cite{liu2024improved} & Vicuna & 7B & 9.9 & 8.4 & 9.1 & 9.5 & 9.5 & 9.5 & 15.0 & 14.0 & 14.5  \\
    LLaVA-FT   & Vicuna & 7B & 22.8 & 18.6 & 20.5 & 14.3 & 10.4 & 12.0 & 24.0 & 20.2 & 21.9 \\
    GPT-4o  \cite{gpt4o2024} & -- & --  & 37.3 & 27.8 &  31.9 & 22.9 & 19.0 & 20.8 & 35.2 & 28.8 & 31.7 \\
    \midrule
    \multicolumn{12}{l}{\textbf{Retrieved Knowledge}} \\
    CLIP→PaLM \cite{chen2023can} & PaLM & 540B & 21.9 & 18.6 & 20.1 & 15.6 & 14.9  & 15.2 & 22.7 & 18.5 & 20.4 \\
    CLIP→FiD  \cite{chen2023can} & T5large & 660M & 20.7 & 18.1 & 19.3 & 18.9 & 17.6 & 18.2 & 23.3 & 19.1 & 20.9 \\
    Wiki-LLaVA  \cite{caffagni2024wiki} & Vicuna & 7B  & -- & -- & -- & -- & -- & -- & 30.1 & 27.8 & 28.9 \\
    EchoSight  \cite{yan2024echosight} & LLaMA3 & 8B  & -- & -- & -- & -- & -- & -- & -- & -- & 31.3 \\
    LLM-RA \cite{jian2024large} & -- & --  & 26.1 & 20.9 & 23.1 & -- & -- & -- & -- & -- & -- \\
    \rowcolor{mygray}
    GPT-4o-mRAG $\dagger$ & -- & -- & 38.9 & 33.0 & 35.7 & 26.8 & 22.6 & 24.5 & 38.1 & 32.1 & 34.8 \\
    \rowcolor{mygray}
    LLaVA-mRAG $\dagger$ & Vicuna & 7B & 30.3 & 29.2 & 29.8 & 17.6 & 15.9 & 16.7 & 30.4  & 31.0 & 30.7 \\
    \rowcolor{mygray}
    LLaVA-SFR $\dagger$ & Vicuna & 7B & 20.8 & 19.1 & 19.9 & 18.5 & 17.2 & 17.9 & 20.8 & 19.7 & 20.3 \\
    \rowcolor{mygray}
    LLaVA-mR$^2$AG $\dagger$ & Vicuna & 7B & \textbf{39.1} & \textbf{38.0} & \textbf{38.6}     & \textbf{30.2} & \textbf{27.5} & \textbf{28.8} & \textbf{40.6} & \textbf{39.8} & \textbf{40.2} \\
    \midrule
    \multicolumn{12}{l}{\textbf{Oracle Knowledge}} \\
    Oracle→FID \cite{chen2023can}  & T5large  & 660M  & --  & -- & 52.0 & -- & -- & 45.6 & 52.1 & 53.0 & 52.5 \\
    Wiki-LLaVA \cite{caffagni2024wiki} & Vicuna & 7B & -- & -- & -- & -- & -- & -- & 52.7  & 50.3 & 51.5  \\
    AVIS \cite{hu2024avis} & -- & -- & 56.4 & 50.7 & 53.4 & -- & -- & -- & -- & -- & -- \\
    \rowcolor{mygray}
    GPT-4o-mRAG $\dagger$ & -- & -- & 56.4 & 54.0 & 55.2 & 44.5 & 39.7 & 42.0 & 60.8  & 54.1 & 57.2 \\
    \rowcolor{mygray}
    LLaVA-mRAG $\dagger$ & Vicuna & 7B & 55.3 & 56.1 & 55.7 & 32.8 & 28.2 & 30.3 & 55.6 & 56.8 &  56.2 \\
    \rowcolor{mygray}
    LLaVA-SFR $\dagger$ & Vicuna & 7B & 56.6 & 55.6 & 56.1 & 46.9 & 43.3 & 45.0 & 52.8 & 53.4 & 53.1 \\
    \rowcolor{mygray}
    LLaVA-mR$^2$AG $\dagger$ & Vicuna & 7B & \textbf{58.3} & \textbf{57.9} & \textbf{58.1} & \textbf{50.4} & \textbf{47.2} & \textbf{48.7} & \textbf{60.8} & \textbf{59.3} & \textbf{60.0} \\
    \bottomrule
\end{tabular}%
}
\end{table*}

\subsection{Datasets}
\textbf{INFOSEEK} \cite{chen2023can} contains a training set and three evaluation sets: INFOSEEK$_{\text{Validation}}$, INFOSEEK$_{\text{Wikidata}}$ and INFOSEEK$_{\text{Human}}$. 
The training set, along with INFOSEEK$_{\text{Validation}}$ and INFOSEEK$_{\text{Wikidata}}$, are all derived from 1.3M samples automatically constructed from Wikipedia to support large-scale training and evaluation. 
INFOSEEK$_{\text{Human}}$ consists of 8.9K samples annotated by humans to simulate real information-seeking intentions.
To prevent overfitting, each evaluation set is divided into two subsets: Unseen Entity and Unseen Question. 
The evaluation samples can be divided into three categories, \ie, \textit{STRING}, \textit{TIME}, and \textit{NUMERICAL}.
The \textit{STRING} and \textit{TIME} categories use VQA Accuracy \cite{goyal2017making} as the evaluation metric, while the \textit{NUMERICAL} category is assessed with Relaxed Accuracy \cite{methani2020plotqa}. 
To calculate the overall accuracy, the average score for each question is first computed separately for each test split,  followed by the geometric mean of these averages.

INFOSEEK uses a Wikipedia knowledge base containing 100K articles and infobox images as external knowledge sources for the With-KB protocol. Since this external knowledge base is not publicly available, we construct one of the same scale and perform comprehensive evaluations across all available datasets. The evaluation metrics strictly follow INFOSEEK's standards, ensuring a fair comparison.

\textbf{Encyclopedic VQA} \cite{mensink2023encyclopedic} contains 1M \{I, Q, A\} triples, covering 16.7K entities and totaling 221K unique Q+A pairs. Each Q+A pair is associated with up to 5 images, showing various instances of the same entity. The dataset includes single-hop questions generated using templated and automatic methods, along with multi-answer and two-hop questions. Multi-answer questions require a list of possible answers, while two-hop questions involve two consecutive retrieval and reasoning steps. For evaluation, accuracy is measured as the percentage of matches between the predicted and ground-truth answers on the test split, using the BERT Matching (BEM) \cite{bulian2022tomayto} standard for correctness assessment. For multi-answer questions, the model's output is first converted into a set of strings, and the intersection-over-the-union (IoU) with the ground-truth set is calculated. If $\text{IoU} \geq 0.5$, the prediction is considered correct; otherwise, BEM is used to determine the equivalence between the predicted and ground-truth lists.

E-VQA provides a controlled knowledge base comprising 2M Wikipedia pages, with each Q+A pair annotated with the corresponding Wikipedia articles and evidence paragraphs supporting the answers. In the retrieval-augmented setting, E-VQA employs a Google Lens-based \cite{google_lens} retriever to predict entities from the input image $I$. We focus on single-hop and multi-answer questions in E-VQA, evaluating on the test split with a consistent knowledge base, retrieval approach and unified evaluation metrics.

\subsection{Implementation details}
\label{sec:implementation_details}

We perform instruction tuning using the combination of the LLaVA  instruction tuning dataset (LLaVA-IT) \cite{liu2024improved} and the mR$^2$AG-IT dataset.
Training continues from the stage-1 checkpoint of the LLaVA-v1.5-7B \cite{liu2024improved}, with a learning rate of \(2 \times 10^{-5}\) and a batch size of \(8 \times 16\), and lasts for one epoch.
The main experiments and ablation studies are conducted based on the LLaVA-v1.5-7B \cite{liu2024improved}, while mR$^2$AG is also applicable to other MLLMs, such as Mini-Gemini \cite{li2024mini} and Mipha \cite{zhu2024mipha}.
We use CLIP-ViT-L/14@336px \cite{radford2021learning} as the retriever for INFOSEEK \cite{chen2023can}, and directly use the retrieval results based on Google Lens for E-VQA \cite{mensink2023encyclopedic}.
By default, we utilize the top-5 Wikipedia entries from the retrieval results for both benchmarks.

\subsection{Comparisons with SOTA}
\label{sec:main_results}

\begin{table}[t!]
\caption{Main results of various methods on E-VQA\cite{mensink2023encyclopedic} benchmark. By default, these methods use Google Lens \cite{google_lens} as the retriever, while methods marked with $^*$ utilize a custom retrieval scheme.}
\centering
\label{tab:enc}
\resizebox{\columnwidth}{!}{%
\begin{tabular}{lcc}
\toprule
\multirow{2}{*}{Model}  & \multicolumn{2}{c}{E-VQA}  \\
                        & Single-hop  & Multi-answer  \\ \midrule
\multicolumn{3}{l}{\textbf{Without Knowledge}}     \\
LLaVA \cite{liu2024improved}         & 14.9\%      & 6.3\%         \\
LLaVA-FT         & 22.8\%      & 18.6\%         \\
GPT-4o \cite{gpt4o2024}      & 27.3\%      & 15.7\%             \\ \midrule
\multicolumn{3}{l}{\textbf{Retrieved Knowledge}}     \\
PaLI \cite{mensink2023encyclopedic}          & 28.1\%      & 9.2\%         \\
PaLM \cite{mensink2023encyclopedic}          & 48.8\%      & 33.6\%        \\
GPT-3 \cite{mensink2023encyclopedic}         & 44.9\%      & 32.1\%        \\
Wiki-LLaVA$^*$ \cite{caffagni2024wiki}       & 21.8\%      & --            \\
EchoSight$^*$ \cite{yan2024echosight}        & 41.8\%      & --            \\
HAMMR-BLIP-2 \cite{castrejon2024hammr}       & 45.0\%      & --            \\
HAMMR-PaLI-X \cite{castrejon2024hammr}       & 47.8\%      & --            \\
Cascade w/o vanilla MLLMs \cite{alazraki2023not}     & 53.4\%      & --    \\
\rowcolor{mygray}
GPT-4o-mRAG   & 49.2\%  & 49.5\%          \\
\rowcolor{mygray}
LLaVA-mR$^2$AG    & \textbf{55.9\%}  & \textbf{51.8\%}          \\ \midrule
\multicolumn{3}{l}{\textbf{Oracle Knowledge}} \\
PaLI \cite{mensink2023encyclopedic}          & 48.8\%      & --        \\
PaLM \cite{mensink2023encyclopedic}          & 87.0\%      & --        \\
GPT-3 \cite{mensink2023encyclopedic}         & 82.1\%      & --        \\
Wiki-LLaVA \cite{caffagni2024wiki}       & 39.2\%          & --        \\
\rowcolor{mygray}
GPT-4o-mRAG       & 87.3\%                 & --                        \\
\rowcolor{mygray}
LLaVA-mR$^2$AG    & \textbf{88.2\%}      & --           \\ \bottomrule
\end{tabular}%
}
\end{table}
\subsubsection{INFOSEEK}

\textbf{Without Knowledge.}
In this setting, the model predicts answers based solely on the input image and question, relying on the knowledge learned during training.
To explore the performance of MLLMs under this protocol, we fine-tune LLaVA-v1.5-7B \cite{liu2024improved} using \{V, Q, A\} triples from the INFOSEEK training set, with results shown in the LLaVA-FT (Fine-Tuned) row of Table~\ref{tab:infoseek}.
After fine-tuning, performance on INFOSEEK$_{\text{Wikidata}}$ increases from 9.1 to 20.5, and on INFOSEEK$_{\text{Human}}$ from 9.5 to 12.0. 
Additionally, the strongest models, GPT-4o \cite{gpt4o2024}, achieves scores of 31.9 on INFOSEEK$_{\text{Wikidata}}$ and 20.8 on INFOSEEK$_{\text{Human}}$. 
These results highlight that while fine-tuning for specific datasets or using stronger models helps improve performance, current models still fall short in knowledge-based VQA tasks, underscoring the need for external knowledge.

\textbf{Retrieved Knowledge.}
This section presents the performance of methods that access external knowledge through retrieval.
By leveraging articles from retrieved Wikipedia entries for answer generation, mR$^2$AG outperforms existing methods across all three test sets.
Specifically, mR$^2$AG surpasses LLM-RA \cite{jian2024large}, CLIP→FiD \cite{chen2023can}, and EchoSight \cite{yan2024echosight} by 15.5\%, 10.6\%, and 8.9\% in overall accuracy on INFOSEEK$_{\text{Wikidata}}$, INFOSEEK$_{\text{Human}}$, and INFOSEEK$_{\text{Validation}}$, respectively.

To confirm that that our improvements stem from the proposed mR$^2$AG framework rather than improved retrieval methods or base MLLM capabilities, we develop two baseline variants: LLaVA-mRAG and LLaVA-SFR. 
LLaVA-mRAG directly utilizes the summary of top-1 retrieved entry, representing the naive mRAG strategy. 
LLaVA-SFR adopts the off-the-shelf model SFR \cite{meng2024sfrembedding}, extracting the most relevant paragraphs from the top-5 retrieved entries, showcasing the performance of employing an external retrieval model to extract relevant information.
Our results significantly outperform both baselines, demonstrating that our method can effectively handles noisy retrieval content, precisely pinpoint the relevant information, and extract the knowledge for answering questions.

We introduce GPT-4o-mRAG as an extended baseline, applying the same augmentation strategy as LLaVA-mRAG to explore the impact of integrating naive mRAG with advanced MLLMs. Results show only marginal performance gains for GPT-4o, for example, accuracy on INFOSEEK$_{\text{Wikidata}}$ improving slightly from 31.9 to 35.7. This highlights the limitations of naive mRAG approaches, which indiscriminately use noisy retrieved content. In contrast, the mR$^2$AG framework’s reflection mechanism proves more effective, underscoring its superiority.

\textbf{Oracle Knowledge.}
Even when ground-truth Wikipedia entries for the entities in the question are available, our method still outperforms others across all test sets. This further highlights the advantage of mR$^2$AG in refining useful evidence and explicitly outputting evaluation results before answer generation, which enhances the model’s ability to perform knowledge reasoning.

\subsubsection{Encyclopedic VQA}
As Table~\ref{tab:enc} illustrates, when using the retrieved knowledge, mR$^2$AG achieves superior performance, improving from 53.4\% to 55.9\% on single-hop questions and significantly increasing from 33.6\% to 51.8\% on the overlooked multi-answer questions. 
It also surpasses both GPT-4o and GPT-4o-mRAG baselines, demonstrating its ability to effectively analyze noisy retrieved knowledge, accurately locate relevant information, and generate high-quality responses.   
Under the Oracle Knowledge setting, with more reliable knowledge sources, mR$^2$AG's performance dramatically improves from 55.9\% to 88.2\%, surpassing the previous state-of-the-art of 87.0\%. 
This highlights two key points:
1) Improving retrieval precision effectively enhances the performance of mR$^2$AG, demonstrating the method’s high potential.
2) With the same retrieval content, mR$^2$AG's superior performance underscores that the Relevance-Reflection mechanism further strengthens the model's information extraction and reasoning capabilities.
Overall, mR$^2$AG sets a new state-of-the-art across multiple Knowledge-based VQA tasks, demonstrating its broad applicability.

\begin{table}[t!]
\caption{Comparisons of different answer post-processing methods.}
\centering
\label{tab:rank}
\footnotesize
\resizebox{\columnwidth}{!}{%
\begin{tabular}{>{\raggedright\arraybackslash}p{1.8cm}>{\centering\arraybackslash}p{2.0cm}>{\centering\arraybackslash}p{1.9cm}c}
\toprule
\multirow{2}{*}{Score Type} & \multicolumn{3}{c}{INFOSEEK$_{\text{Validation}}$}     \\
                            & Unseen Question & Unseen Entity & Overall              \\ \midrule
Random & 32.7 & 32.7 & 32.7 \\
$S^{\mathrm{ans}}$ & 36.4 & 37.0 & 36.7 \\
$S^{\mathrm{ret}}$ & 37.6 & 37.4 & 37.5 \\
$S^{\mathrm{rel}}$ & 39.7 & 38.7 & 39.2 \\
$S^{\mathrm{ret}} \cdot S^{\mathrm{ans}}$ & 38.3 & 38.7 & 38.5 \\
$S^{\mathrm{rel}} \cdot S^{\mathrm{ans}}$ & 39.5 & 39.0 & 39.2 \\
$S^{\mathrm{ret}} \cdot S^{\mathrm{rel}}$ & 40.5 & 39.3 & 39.9 \\
$S^{\mathrm{ret}} \cdot S^{\mathrm{rel}} \cdot S^{\mathrm{ans}}$ & \textbf{40.6} & \textbf{39.8} & \textbf{40.2} \\ \bottomrule
\end{tabular}%
}
\end{table}
\begin{table}[t!]
\caption{Effect of retrieving different numbers of Wikipedia entries.}
\centering
\tiny
\label{tab:topk}
\resizebox{\columnwidth}{!}{%
\begin{tabular}{l>{\centering\arraybackslash}p{0.5cm}>{\centering\arraybackslash}p{0.5cm}>{\centering\arraybackslash}p{0.5cm}>{\centering\arraybackslash}p{0.5cm}}
\toprule 
Retrieved Entries & 1  & 5 & 10  & 20   \\ \midrule
INFOSEEK$_{\text{Validation}}$  & 29.7  & 40.2 & 40.5  & 40.4  \\
R@K & 38.0  & 58.6  & 65.0 & 70.6  \\ \bottomrule
\end{tabular}%
}
\end{table}

\subsection{Ablation Studies}

In this section, we conduct ablation studies to verify the effectiveness of our model's design choices. These comparisons are conducted on the INFOSEEK$_{\text{Validation}}$ dataset by default.

\textbf{Effect of $S^{\mathrm{ret}}$, $S^{\mathrm{rel}}$ and $S^{\mathrm{ans}}$.}
To evaluate the effectiveness of using the product of the Retrieval-Reflection score $S^{\mathrm{ret}}$, Relevance-Reflection score $S^{\mathrm{rel}}$, and answer confidence score $S^{\mathrm{ans}}$ in post-processing, we conducted a comprehensive comparison with all alternative ranking methods, as shown in Table~\ref{tab:rank}. As observed, randomly selecting from answer candidates yields the lowest performance, which is not desirable. When using scores for post-processing, the combination of all three scores is superior to other score combinations, demonstrating that these scores effectively assess the reliability of the answer at different levels.

\textbf{Effect of retrieving different numbers of Wikipedia entries.}
To provide insights into the optimal number of retrieved Wikipedia entries for augmenting generation, we explore how varying the number of entries affects performance, as shown in Table~\ref{tab:topk}.
Increasing the number of entries from 1 to 5 significantly improves the recall rate (+20.6) and leads to a notable increase in performance on INFOSEEK$_{\text{Validation}}$ (+10.5), indicating that this process effectively enhances the likelihood of capturing relevant information.
However, as the number of entries continues to rise, the improvements become less significant. 
At 10 entries, the accuracy shows only a slight increase, and at 20 entries, performance even declines.
This suggests that too many retrieved entries introduce excessive noise, limiting further gains.
Considering the trade-off between performance and efficiency, we select 5 retrieved Wikipedia entries for each question.

\textbf{Contribution of the NQ samples to the mR$^2$AG-IT dataset.}
NQ samples serve as a supplement to the mR$^2$AG-IT dataset, which contains samples with evidence paragraphs. As shown in Table~\ref{tab:w/o_nq}, removing this portion of data leads to a performance decline. This finding highlights the importance of including NQ data in enhancing the model's ability to accurately identify evidence paragraphs.

\begin{table}[t!]
\caption{The importance of NQ samples.}
\centering
\footnotesize
\label{tab:w/o_nq}
\resizebox{\columnwidth}{!}{%
\begin{tabular}{>{\centering\arraybackslash}p{1.2cm}>{\centering\arraybackslash}p{2.2cm}>{\centering\arraybackslash}p{2.2cm}>{\centering\arraybackslash}p{1.0cm}}
\toprule
 & \multicolumn{3}{c}{INFOSEEK$_{\text{Validation}}$}  \\
 & Unseen Question & Unseen Entity & Overall \\ \midrule
w/o NQ & 39.1 & 39.7 & 39.4 \\
w/ NQ & \textbf{40.6} & \textbf{39.8} & \textbf{40.2} \\ \bottomrule
\end{tabular}%
}
\end{table}

\textbf{Effect of matched modality in retrieval.}
Each Wikipedia entry contains a candidate image and a title. We compare three retrieval configurations that match the query image against different entry modalities via CLIP: the entry image only (Image), the entry title only (Title), and both (Title + Image). Table~\ref{tab:entity_recall} reports the retrieval recall for locating ground-truth Wikipedia entities on INFOSEEK$_{\text{Validation}}$. The combined Title + Image strategy significantly outperforms both single-modality variants across all recall metrics (e.g., R@5: 58.8 vs.\ 51.7/48.0). Table~\ref{tab:retrieval_vqa_ablation} further reports the downstream VQA accuracy when each configuration provides the top-5 evidence articles. The same trend carries over: Title + Image achieves the best overall accuracy (40.2 vs.\ 36.8/34.9).

\begin{table}[t!]
\caption{Retrieval recall on the INFOSEEK knowledge base. Matched Modality indicates the Wikipedia entry modality used for CLIP similarity matching with the query image.}
\centering
\tiny
\label{tab:entity_recall}
\resizebox{\columnwidth}{!}{%
\begin{tabular}{lcccc}
\toprule
Matched Modality & R@1 & R@5 & R@10 & R@20 \\ \midrule
Title & 31.2 & 51.7 & 58.7 & 65.1 \\
Image & 29.1 & 48.0 & 54.8 & 62.0 \\
Title + Image & \textbf{38.0} & \textbf{58.8} & \textbf{65.0} & \textbf{70.6} \\ \bottomrule
\end{tabular}%
}
\end{table}

\begin{table}[t!]
\caption{Effect of matched modality on the downstream accuracy of INFOSEEK$_{\text{Validation}}$.}
\centering
\footnotesize
\label{tab:retrieval_vqa_ablation}
\resizebox{\columnwidth}{!}{%
\begin{tabular}{lcccc}
\toprule
Matched Modality & R@5 & Unseen Question & Unseen Entity & Overall \\ \midrule
Title & 51.7 & 37.2 & 36.4 & 36.8 \\
Image & 48.0 & 34.6 & 35.2 & 34.9 \\
Title + Image & \textbf{58.8} & \textbf{40.6} & \textbf{39.8} & \textbf{40.2} \\ \bottomrule
\end{tabular}%
}
\end{table}

\begin{table}[t!]
\caption{Results on MLLMs with different architectures and scales.}
\centering
\normalsize
\label{tab:gen_sca}
\resizebox{\columnwidth}{!}{%
\begin{tabular}{>{\raggedright\arraybackslash}p{3.0cm}lr>{\centering\arraybackslash}p{1.7cm}>{\centering\arraybackslash}p{1.7cm}}
\toprule
Model       & LLM        & \#Params & \begin{tabular}[c]{@{}c@{}}INFOSEEK \\ Wikidata \end{tabular} & \begin{tabular}[c]{@{}c@{}}INFOSEEK \\ Human \end{tabular} \\ \midrule
\multicolumn{5}{l}{\textbf{Without Knowledge}}                                \\
Mipha \cite{zhu2024mipha}         & Phi2    & 3B       & 7.0              & 6.4           \\
MGM   \cite{li2024mini}        & Vicuna  & 7B       & 10.7             & 10.2          \\
LLaVA  \cite{liu2024improved}        & Vicuna  & 13B      & 9.8              & 9.9           \\ \midrule
\multicolumn{5}{l}{\textbf{Retrieved Knowledge}}                              \\
Mipha-mRAG      & Phi2    & 3B       & 26.6             & 14.4          \\
MGM-mRAG        & Vicuna  & 7B       & 29.8             & 18.4          \\
LLaVA-mRAG      & Vicuna  & 13B      & 31.3             & 18.1          \\ \rowcolor{mygray}
Mipha-mR$^2$AG & Phi2    & 3B       & \textbf{34.4}             & \textbf{26.3}          \\ \rowcolor{mygray} 
MGM-mR$^2$AG   & Vicuna  & 7B       & \textbf{38.4}             & \textbf{28.8}          \\ \rowcolor{mygray}
LLaVA-mR$^2$AG & Vicuna  & 13B      & \textbf{39.7}             & \textbf{29.4}          \\ \bottomrule
\end{tabular}%
}
\end{table}

\subsection{Generalizability and Scalability}
We validate the effectiveness of the mR$^2$AG framework on three model architectures (Mipha \cite{zhu2024mipha}, Mini-Gemini \cite{li2024mini}, and LLaVA \cite{liu2024improved}), covering three different scales of language models (3B, 7B, and 13B). In our experiments, we fine-tune these MLLMs from their stage-1 checkpoints using the same instruction-tuning data, while maintaining consistent hyperparameters such as learning rate, batch size and number of epochs. The results in Table~\ref{tab:gen_sca} show that these base models generally perform poorly on the INFOSEEK \cite{chen2023can} benchmark. However, by introducing the naive mRAG approach or the mR$^2$AG framework, we observe significant improvements in performance on Knowledge-based VQA tasks. Notably, the mR$^2$AG framework consistently outperforms the naive mRAG approach, demonstrating its generalizability  across different model architectures and scalability across varying language model sizes.

\begin{table*}[t!]
\caption{Main results on standard VQA benchmarks. LLaVA-Wild denotes LLaVA-Bench (In-the-Wild)~\cite{liu2023visual}.}
\centering
\normalsize
\label{tab:mllm_bench}
\resizebox{\textwidth}{!}{%
\begin{tabular}{lcccccccc}
\toprule
\multirow{2.5}{*}{Model} & \multicolumn{2}{c}{MME\cite{fu2024mme}} & \multirow{2.5}{*}{LLaVA-Wild\cite{liu2023visual}} & \multirow{2.5}{*}{MMBench\cite{liu2024mmbench}} & \multirow{2.5}{*}{POPE\cite{li2023evaluating}} & \multirow{2.5}{*}{VQAv2\cite{goyal2017making}} & \multirow{2.5}{*}{GQA\cite{hudson2019gqa}} & \multirow{2.5}{*}{OK-VQA\cite{marino2019ok}} \\ \cmidrule(lr){2-3}
& Cognition & Perception &  &  &  &  &  &  \\ \midrule
LLaVA-v1.5-7B \cite{liu2024improved} & 311.4 & 1490.4 & 64.1 & 65.8 & 85.6 & 78.2 & 62.0 & 56.6 \\
\rowcolor{mygray}
LLaVA-mR$^2$AG & 325.7 & 1476.2 & 65.8 & 66.2 & 86.1 & 78.3 & 61.8 & 57.1 \\ \bottomrule
\end{tabular}%
}
\end{table*}

\subsection{Generalizability on Standard VQA Benchmarks}
\label{sec:llava_bench}
We evaluate the generalizability of mR$^2$AG on standard VQA benchmarks to verify that knowledge-oriented instruction tuning does not degrade performance on other tasks. We reproduce the base model (LLaVA-v1.5-7B\cite{liu2024improved}) under identical training conditions and conduct a comprehensive evaluation, as shown in Table~\ref{tab:mllm_bench}.
Across all evaluated benchmarks, mR$^2$AG maintains comparable or improved performance relative to the base model. Specifically, mR$^2$AG achieves comparable scores on MME\cite{fu2024mme}, VQAv2\cite{goyal2017making}, and GQA\cite{hudson2019gqa}, while showing slight improvements on LLaVA-Wild\cite{liu2023visual}, MMBench\cite{liu2024mmbench}, and POPE\cite{li2023evaluating}, confirming that RAG-oriented instruction tuning preserves the base model's visual understanding capabilities. In particular, on OK-VQA\cite{marino2019ok}, which requires commonsense knowledge, mR$^2$AG achieves a slight improvement over the base model (57.1 vs.\ 56.6, +0.5), suggesting that knowledge-oriented instruction tuning can also benefit tasks involving commonsense reasoning.
These results demonstrate that mR$^2$AG achieves targeted optimization for knowledge-based VQA while maintaining strong generalizability on standard VQA benchmarks.

\begin{figure*}[t!]
  \centering
  \includegraphics[width=\textwidth]{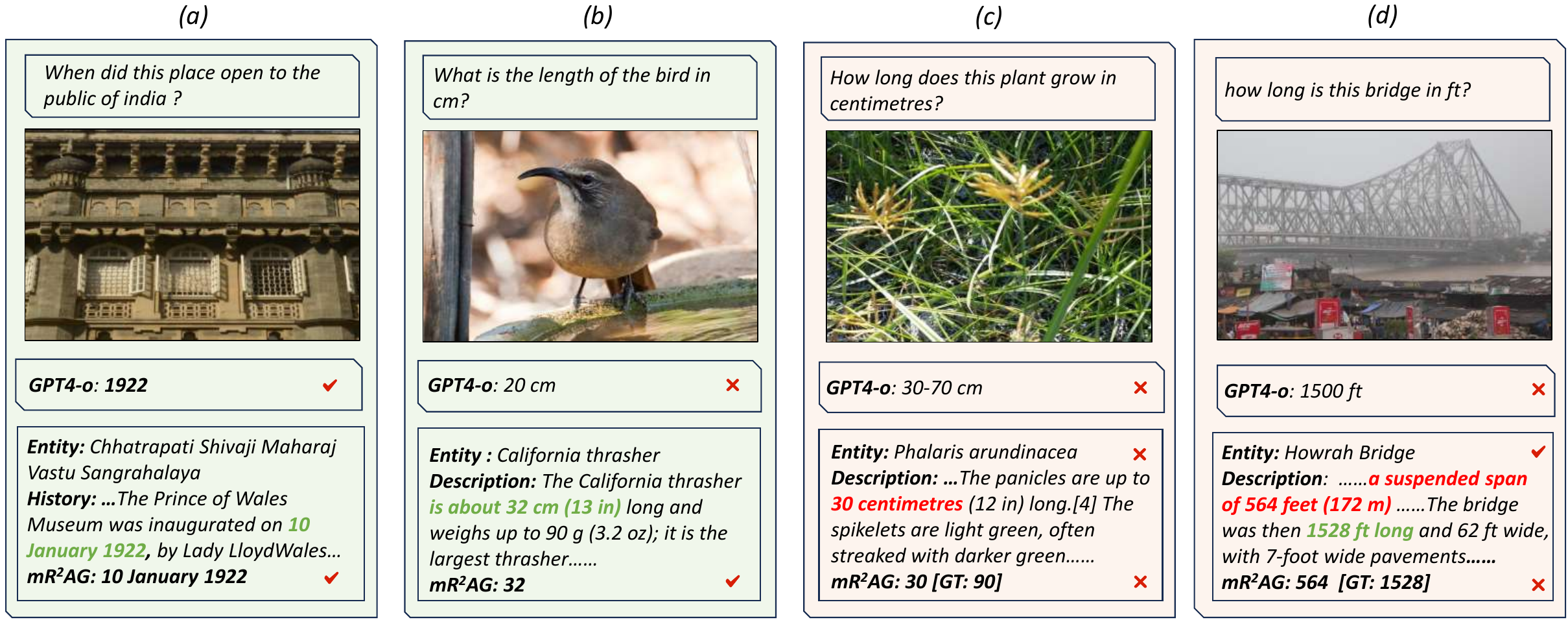}
  \caption{Qualitative comparison of GPT-4o and mR$^2$AG on INFOSEEK dataset. Two failure cases are shown in the (c) and (d).}
  \label{fig:3}
\end{figure*}

\begin{table}[t!]
\caption{Comparison of inference time and performance across methods.}
\label{tab:inference}
\centering
\footnotesize
\resizebox{\columnwidth}{!}{%
\begin{tabular}{llccc}
\toprule
Method & Evidence & INFOSEEK$_{\text{Val}}$ & Time (s) & \# Calls \\ \midrule
LLaVA-FT & --- & 21.9 & 0.26 & 0 \\
LLaVA-mRAG & 1 entry, summary & 30.7 & 0.32 & 1 \\
LLaVA-mR$^2$AG & 1 entry, summary & -- & 0.34 & 1 \\
LLaVA-mRAG & 1 entry, full article & 31.8 & 0.52 & 1 \\
LLaVA-mR$^2$AG & 5 entries, full articles & 37.2 & 2.83 & 5 \\
LLaVA-mR$^2$AG & 5 entries, paragraphs & \textbf{40.2} & 3.03 & 39.4 \\
\bottomrule
\end{tabular}%
}
\end{table}

\subsection{Computational Efficiency and Performance Trade-offs}
\label{sec:efficiency}
Table~\ref{tab:inference} quantifies the computational trade-offs across different methods on the INFOSEEK$_{\text{Validation}}$ set, measuring both average response time per query and number of model calls per query.
We further implement a FullArticle baseline that directly feeds the full text of each top-5 retrieved entry without paragraph-level filtering, to analyze the impact of input context length on both performance and computational efficiency.

\begin{itemize}[leftmargin=*]
\item The reflection mechanism introduces negligible computational overhead. When processing the same retrieved content, LLaVA-mR$^2$AG requires only 0.34 seconds compared to 0.32 seconds for LLaVA-mRAG without reflection tokens, as the mechanism only generates a single additional token during inference.
\item Directly increasing input context length raises inference time with only limited performance gains. For LLaVA-mRAG on the same top-1 entry, switching from summary to full article increases inference time from 0.32 to 0.52 seconds, while accuracy improves only marginally (30.7 $\to$ 31.8). At the top-5 retrieval scale, although the FullArticle baseline requires fewer model calls than paragraph-level Relevance-Reflection, the total inference time remains comparable (2.83 vs.\ 3.03s); moreover, the longer context increases the difficulty of evidence localization, resulting in significantly lower accuracy (37.2) compared to the latter (40.2, +3.0).
\item LLaVA-mR$^2$AG employs two targeted optimization strategies: 1) Parallel inference: processing multiple paragraphs in batches; and 2) Dynamic pruning: for irrelevant paragraphs, the model returns an [Irrelevant] token and immediately terminates the generation process, filtering out irrelevant content while minimizing computational overhead, thereby efficiently handling multiple retrieved paragraphs in practical inference.
\end{itemize}
These optimizations effectively prevent linear time growth (3.03/0.34) with increasing paragraphs (39.4/1), enabling paragraph-level Relevance-Reflection to achieve a better trade-off between answer accuracy and computational efficiency.

\begin{table}[t!]
\caption{Accuracy of Reflection tokens generation across benchmarks. In E-VQA, S and M refer to Single-hop and Multi-answer, respectively. INFOSEEK$^\text{V}$ refers to INFOSEEK$_{\text{Validation}}$.}
\label{tab:ref_acc}
\centering
\Large
\resizebox{\columnwidth}{!}{%
\begin{tabular}{lccccccc}
\toprule
\multirow{2}{*}{Type}  & \multicolumn{2}{c}{E-VQA} & \multirow{2}{*}{ INFOSEEK$^\text{V}$} & \multirow{2}{*}{MME} & \multirow{2}{*}{LLaVA-Wild} & \multirow{2}{*}{MMBench} & \multirow{2}{*}{POPE} \\ \cmidrule(lr){2-3}
  & S & M &  &  &  &  &  \\ \midrule
Retrieval  & 99.7 & 99.8  & 99.2  & 98.9  & 96.7  & 100 & 100 \\
Relevance  & 93.9 & 94.9  & 91.7  & -     & -     & -   & -  \\ \bottomrule
\end{tabular}%
}
\end{table}

\subsection{Effectiveness of Reflection Tokens}
In Table~\ref{tab:ref_acc}, we independently validate the effectiveness of each reflection token:
\begin{itemize}[leftmargin=*]
    \item \textbf{Retrieval-Reflection:} Based on the Visual-dependent and Knowledge-based benchmark types, we set the Retrieval-Reflection types as [No Retrieval] and [Retrieval], respectively. Results show error rates below 3.3\% and 0.8\%, confirming the validity of this token design.
    \item \textbf{Relevance-Reflection:} For the E-VQA dataset, we determine the Relevance-Reflection token based on the annotated evidence section. In the INFOSEEK$_{\text{Validation}}$ set, the type of Relevance-Reflection token for each paragraph is annotated by GPT-4 \cite{achiam2023gpt}. Under Retrieved Knowledge, the Relevance-Reflection token achieves over 91\% accuracy, demonstrating its effectiveness.
\end{itemize}

\subsection{Qualitative Analysis}
In Fig.~\ref{fig:3}, we present the qualitative comparison of the mR$^2$AG with GPT-4o. As illustrated in Fig.~\ref{fig:3} (a) and (b), our method retrieves relevant knowledge and accurately answers the Knowledge-based questions, achieving more precise results compared to GPT-4o.
However, two failure cases are shown in Fig.~\ref{fig:3}, explained by the following:
\begin{itemize}[leftmargin=*]
    \item Inaccurate retrieval: As illustrated in Fig.~\ref{fig:3} (c), when the subject in the image is difficult to identify, the retriever struggles to find relevant information, making it challenging for our method to answer the questions.
    \item Knowledge interference: In Fig.~\ref{fig:3} (d), the retriever finds the correct entity, but our method provides the wrong answer due to the conflicting knowledge in the text, specifically ''a suspended span of 564 feet.''
\end{itemize}

\section{Conclusion}
\label{sec:conclusion}
This paper proposes an advanced multimodal RAG framework that optimizes Knowledge-based VQA tasks while maintaining the model's capabilities as a general-purpose MLLM. The approach is based on existing MLLMs, guiding the model to explicitly distinguish the type of user query and evaluate retrieved information to refine the naive multimodal RAG process. It improves inference efficiency by adaptively avoiding unnecessary retrievals. Furthermore, by explicitly evaluating the retrieved content, it can identify the evidence passages relevant to the query, while filtering out noise from the retrieved content, thereby enhancing the credibility of the generated responses. Future work will explore knowledge graph-based retrieval-augmented systems and broader application scenarios.

\section{Limitation}
\label{sec:limitation}
Limited by the entity recognition capabilities of existing MLLMs, our method is quite dependent on the retriever, assuming that the default visual entity occupies the major position in the image. When the correct target entity is entirely absent from the retrieval results, the accuracy of the final answer can be affected. To this end, we plan to explore more effective information filtering and extraction methods to broaden the retrieval coverage, incorporate a fallback mechanism that allows the model to abstain from answering or proactively conduct broader retrieval when the retrieved evidence is deemed unreliable, and conduct specialized training for visual entity recognition to handle diverse and complex scenes.

\section*{Acknowledgments}
This research was supported by multiple funding sources, including the New Generation Artificial Intelligence-National Science and Technology Major Project (2025ZD0123401), the National Natural Science Foundation of China (U24A20331, 62302501, 62192782, 62532015, 62536002, U2441241), Beijing Natural Science Foundation (L251005, L223003, L243015, L252032), and Beijing Major Science and Technology Project (Z251100008425008).

{
    \bibliographystyle{IEEEtran}
    \bibliography{main}
}

{\appendices
\section*{Supplementary Material}
\subsection{Prompt Engineering}
\subsubsection{\texorpdfstring{mR$^2$AG-IT Dataset Annotation}{mR2AG-IT Dataset Annotation}}
We utilize the GPT-4 \cite{achiam2023gpt} model via API to annotate the training dataset and design the following prompt to assess the relevance between retrieved content and the query.
Inspired by the chain-of-thought \cite{wei2022chain} approach, the prompt instructs GPT-4 \cite{achiam2023gpt}
to extract evidence sentences before generating relevance judgments.
This design not only enhances the accuracy of the judgments but also facilitates manual calibration.
For each input, the content within curly braces \{\} is replaced with the corresponding actual input.
In the following, ''question'', ''answer'', and ''paragraph'' correspond to $Q$, $A$, and $P_{ij}$, respectively.

\begin{figure*}
  \centering
  \includegraphics[width=0.99\textwidth]{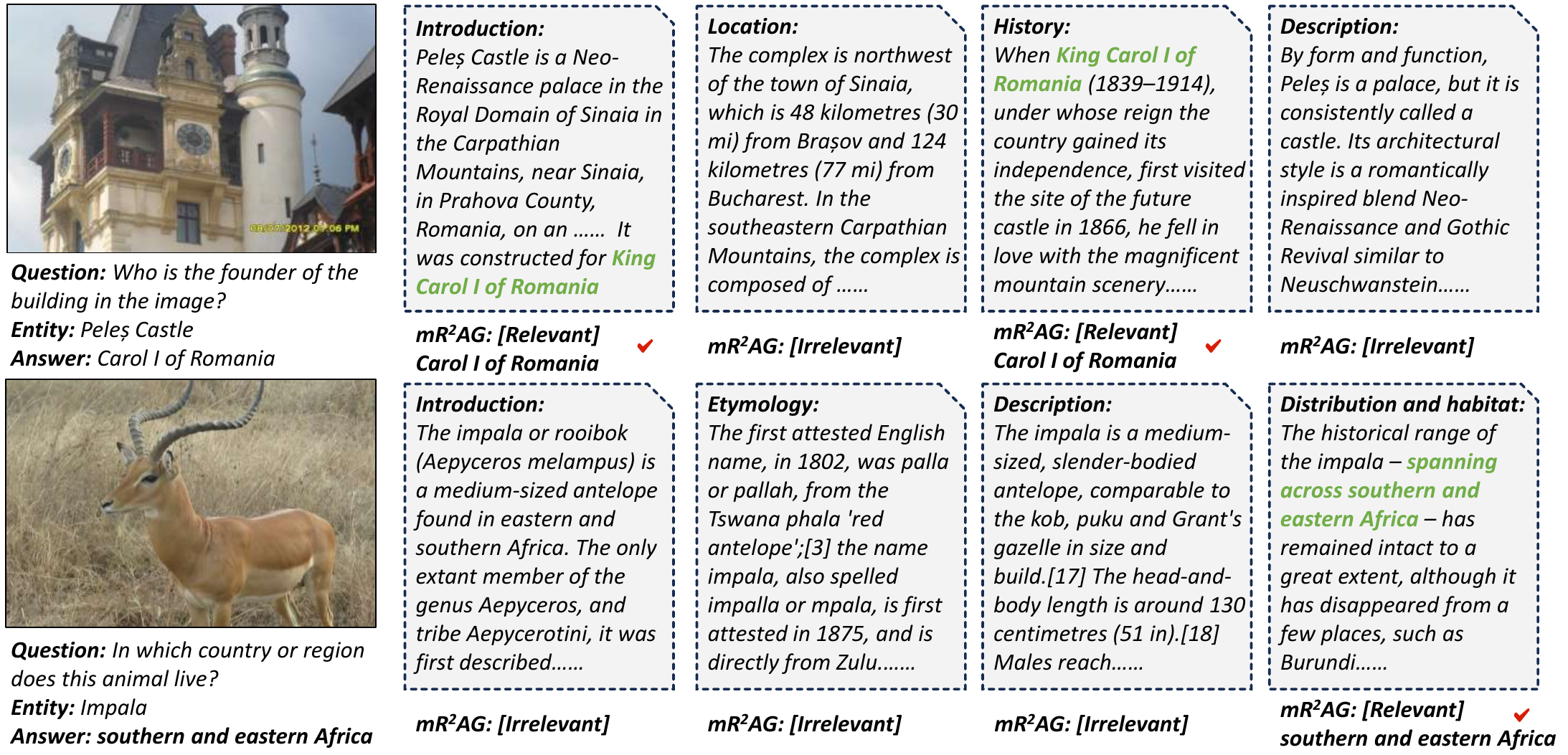}
  \caption{Qualitative results showing the effectiveness of the mR$^2$AG framework. The first row shows results from the INFOSEEK \cite{chen2023can} dataset, while the second row shows results from E-VQA \cite{mensink2023encyclopedic}.}
  \label{fig:4}
\end{figure*}

\begin{figure*}
  \centering
  \includegraphics[width=\textwidth]{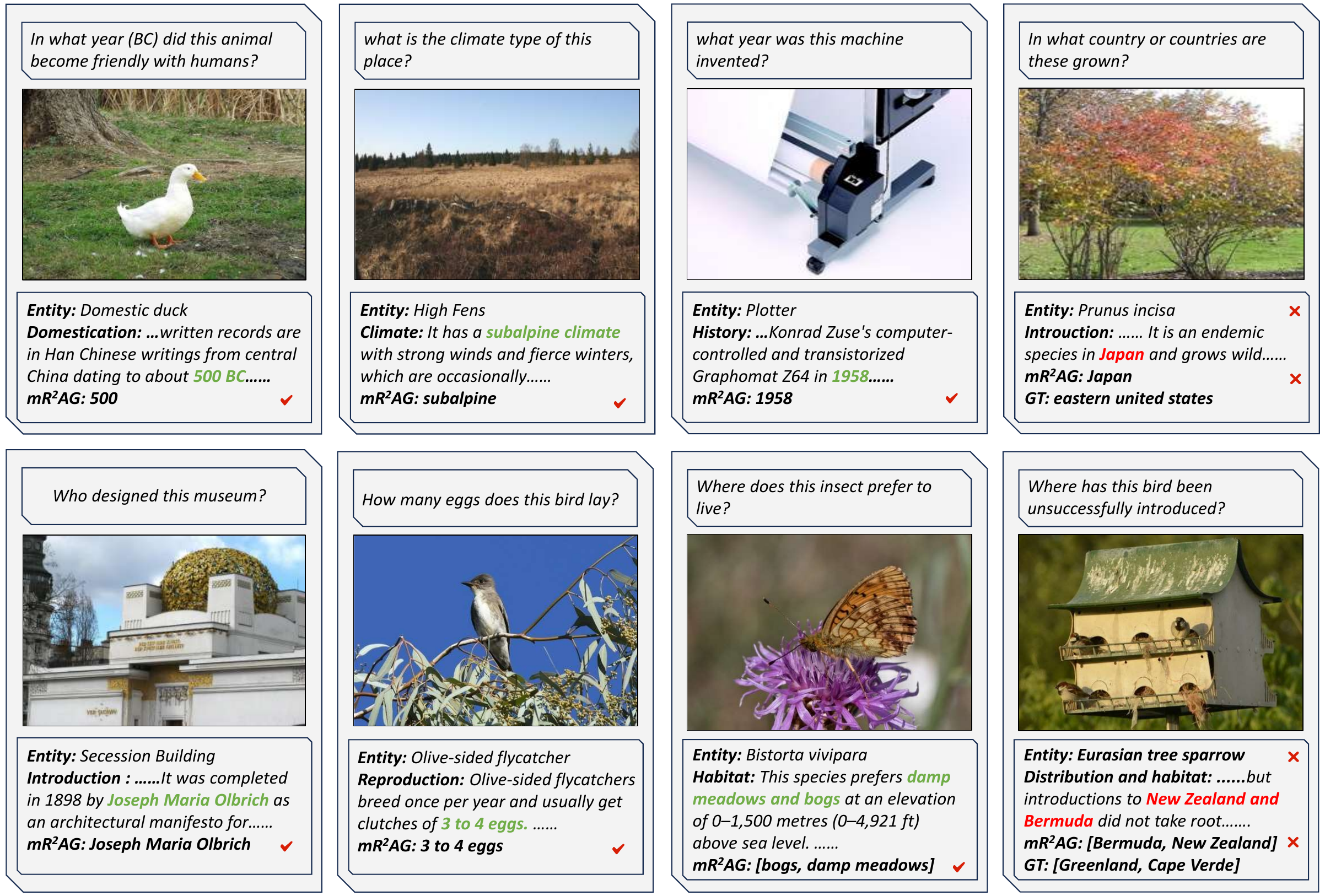}
  \caption{Additional visualization results are provided: the first row shows examples from INFOSEEK \cite{chen2023can}; the second row shows examples from E-VQA \cite{mensink2023encyclopedic}, covering single-hop and multi-answer questions. The last column presents incorrect answers.}
  \label{fig:5}
\end{figure*}

\begin{tcolorbox}[colback=gray!10, colframe=black, title=Dataset Annotation Prompt for GPT-4]
\textbf{Instruction:}

Given a question and its corresponding answer, I need your help to verify whether the retrieved document provided below can fully and effectively support the corresponding answer to the question, and then accurately locate the source of the answer within the paragraph. If so, please respond with \textbf{[Relevant]} and find the evidence sentence supporting the answer. If not, please just respond with \textbf{[Irrelevant]}.

There are only two formats for your response:
\begin{enumerate}
    \item \textnormal{\textbf{[Relevant]}} \\
    Answer source: source sentence.
    \item \textnormal{\textbf{[Irrelevant]}}
\end{enumerate}

\textbf{Input:}

Question: \{\texttt{question}\}.

Answer: \{\texttt{answer}\}.

Retrieved document: \{\texttt{paragraph}\}.
\end{tcolorbox}

\subsubsection{INFOSEEK}

INFOSEEK \cite{chen2023can} evaluates generated answers using exact match, requiring the outputs to strictly match the annotated answers, which are typically concise and presented in the form of a single word or phrase.
To ensure the outputs align with these requirements, we design the following prompt, guiding the model to focus on retrieved content and produce concise responses:

\textit{''Based on the retrieved document, answer the question with a single word or phrase.''}

\subsubsection{Encyclopedic-VQA}
The E-VQA \cite{mensink2023encyclopedic} dataset includes both single-hop and multi-answer questions. For single-hop questions, we adopt the same prompt template as INFOSEEK. For multiple-answer questions, the model needs to generate several possible answers. 
We adjust the prompt to ensure the answers comply with the dataset requirements, 
enabling effective extraction of answer lists from the responses for evaluation:

\textit{''Based on the retrieved documents, answer the question as briefly as possible, using '\&\&' to connect multiple different answers.''}

\subsection{Qualitative Results and Visualizations}
Fig.~\ref{fig:4} qualitatively demonstrates the effectiveness of the mR$^2$AG framework.
It highlights the framework's ability to accurately assess the relevance between retrieved content and user queries, precisely locate evidence paragraphs within the retrieved documents, and generate reliable answers. 
Fig.~\ref{fig:5} provides additional visualization results, illustrating that mR$^2$AG effectively handles various types of visual entities and question types, further validating the design's effectiveness and reliability. The last column presents additional error cases, where the primary issue lies in the failure to retrieve relevant Wikipedia entities for the visual content.
}

\end{document}